\documentclass{vgtc}                          

\ifpdf
  \pdfoutput=1\relax                   
  \usepackage{graphicx}                
  \DeclareGraphicsExtensions{.pdf,.png,.jpg,.jpeg} 
\else
  \usepackage{graphicx}                
  \DeclareGraphicsExtensions{.eps}     
\fi%

\graphicspath{{figures/}{pictures/}{images/}{./}} 

\usepackage{microtype}                 
\PassOptionsToPackage{warn}{textcomp}  
\usepackage{textcomp}                  
\usepackage{mathptmx}                  
\usepackage{times}                     
\usepackage{cite}                      
\usepackage{tabu}                      
\usepackage{booktabs}                  

\usepackage{framed,multirow}

\usepackage{amssymb}
\usepackage{latexsym}

\usepackage{url}

\usepackage{hyperref}

\usepackage{preamble/beimath}
\usepackage{preamble/textoolz}
\usepackage{subfig}
\usepackage{comment}

\onlineid{1016}

\vgtccategory{Research}

\vgtcinsertpkg

\title{Multi-field Visualisation via Trait-induced Merge Trees}

\author{Jochen Jankowai\thanks{e-mail: jochen.jankowai@liu.se}\\ %
        \scriptsize Linköping University %
\and Talha Bin Masood\thanks{e-mail: talha.bin.masood@liu.se }\\ %
     \scriptsize Linköping University %
\and Ingrid Hotz\thanks{e-mail: ingrid.hotz@liu.se}\\ %
     \parbox{1.4in}{\scriptsize \centering Linköping University}}

\teaser{
  \centering
  \includegraphics[width=\linewidth]{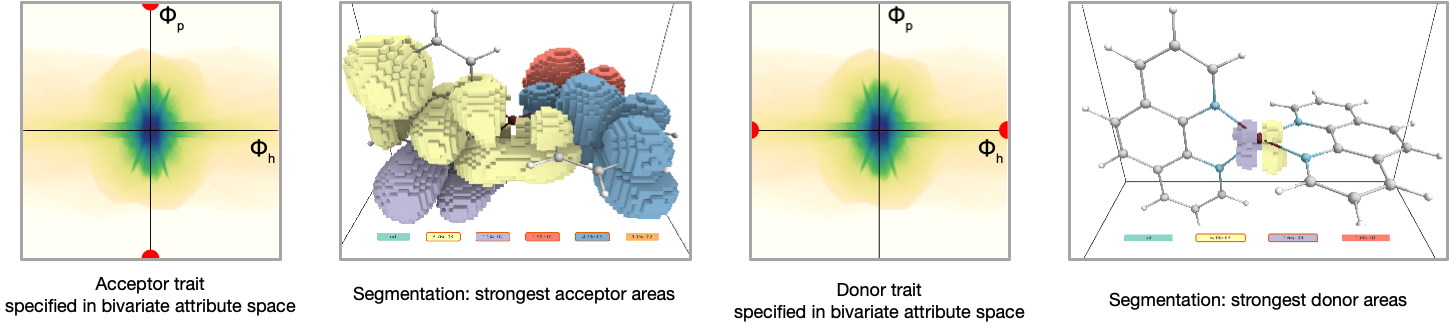}
  \caption{The charge transition of molecules can be described as a bivariate density field ($\Phi_h,\Phi_p$). The image shows two different traits highlighting acceptor respective donor behavior of the molecule in a spatial segmentation. The traits are specified by a pair of points with high absolute values in $\Phi_p$, respective $\Phi_h$.}
  \label{fig:teaser}
}

\abstract{
In this work, we propose trait-based merge trees a generalization of merge trees to feature level sets, targeting the analysis of tensor field or general multi-variate data. For this, we employ the notion of traits defined in attribute space as introduced in the feature level sets framework. The resulting distance field in attribute space induces a scalar field in the spatial domain that serves as input for topological data analysis. The leaves in the merge tree represent those areas in the input data that are closest to the defined trait and thus most closely resemble the defined feature. Hence, the merge tree yields a hierarchy of features that allows for querying the most relevant and persistent features. The presented method includes different query methods for the tree which enable the highlighting of different aspects. We demonstrate the cross-application capabilities of this approach with three case studies from different domains.
} 

\CCScatlist{
  \CCScatTwelve{Human-centered computing}{Visu\-al\-iza\-tion}{Visualization design and evaluation methods}{};
  \CCScatTwelve{Human-centered computing}{Visu\-al\-iza\-tion}{Visualization application domains}{Scientific visualization}
}

\begin{document}


\firstsection{Introduction}

\maketitle


\firstsection{Introduction}
\label{sec:introduction}
Typically, natural phenomena are complex and their simulations need to fit this complexity in the model and output. Oftentimes, this results in a multitude of output variables. Examples include meteorology data containing fields for pressure, temperature, and precipitation, data from computational chemistry simulations often yield a set of density fields, or simulations within mechanical engineering usually include tensor data whose scalar derivatives may be interpreted as multi-variate data.
The visualization of multi-field data has, however, long been an enigma. Methods for scalar data may be applied to each field individually but usually do not extend to a joint analysis of all fields. 
A typical solution is to simultaneously visualize several fields in a display matrix, which means a heavy cognitive load for the users as they need to combine the separate images mentally into one. Including interactive methods in a space of reduced dimensionality can alleviate this mental strain only partially. 

The situation is similar when it comes to topological data analysis (TDA).
TDA has become a fundamental analysis tool for scalar fields in scientific visualization due to its great potential for data abstraction and aggregation~\cite{topologystar}. A large number of typological descriptors exist that are used to generate automatic visualizations or guide interactive exploration. In contrast, the use of topological methods in multi-field visualization is still a largely unexplored domain~\cite{Yan2021b}. There are some interesting ideas to extend concepts from TDA to multi-variate fields. Examples are Reeb spaces a high-dimensional analog of Reeb graphs~\cite{ReebSpacesofPiecewiseLinearMappings}, joint contour nets (JCN)~\cite{jointcontournets} a discrete approximation of the Reeb space, or Jacobi sets, a concept that highlight points where the gradients of the individual scalar functions align~\cite{JacobiSetsofMultipleMorseFunctions}.
However many of the proposed solutions involve complex theoretical concepts that are still challenging to use in practical applications.
Recent advancements address these issues by opening up key scalar field visualization methods, such as iso-surface and volume rendering~\cite{Jankowai2020a}, to multi-variate data. Namely, fiber surfaces and feature level sets (FLS)~\cite{JankowaiHotz2019}. Initially, fiber surfaces have been introduced by Carr et al.~\cite{fibersurfaces} as a method for bi-variate data and have later been extended by Blecha et al.~\cite{raithsalamislice2} to general multi-variate data. Fiber surfaces rely on the intersection of mesh cells and user-given polygons in attribute space. If there is no intersection, the method will not yield a fiber surface. FLS remedies this restriction by treating the case of intersection as the zero level-set of a distance field. The distance field is computed with respect to a geometry and/or range in attribute space. Similar to fiber surfaces, this geometry is defined by the user. It is called a \emph{trait} and defines the parameter configurations the user wishes to find in the data. As stated earlier, the zero level-set of this distance field will yield the same surface as a fiber surface. However, if the zero level set is empty, the user is able to investigate how close the field gets to the trait by rendering different level sets of the distance field.
With the feature definition via traits and their extraction in place, a useful next step for multi-field visualization is a topological analysis of feature level sets. Even though a trait defines and extracts a parameter setting of interest, little is known about the neighborhood of features and the overall structure of the resulting distance field. Depending on the problem statement, the occurrence of highly local features vs. larger features might be of interest. Noise may generate data points fulfilling the trait criteria and clutter the distance field. Topological analysis, in particular the use of merge trees, addresses these concerns in a robust manner. 
The proposed method makes the features extracted through FLS computation browsable through an interface to interact with the corresponding merge tree. The combination of FLS and scalar field topology provides a simple concept for topological analysis of multi-variate fields.
%
%
\paragraph{Contributions} Our approach combines two straightforward and established methods which, in combination, allow for topological analysis of multi-variate data and tensor fields.   We first compute a distance field using the FLS method and use it as input to compute a merge tree, see Figure~\ref{fig:pipline}. The choice of merge trees comes naturally as the leaves in the tree correspond to those areas in the data that are closest to the trait(s). The trait-induced merge tree opens up an array of topology-based simplification and query methods that can be used to gain further insight into the underlying structure of the data. The method is flexible with respect to chosen features of interest represented by traits in attribute space and the results are easy to interpret.

\section{Related work}
\label{sec:relatedwork}
The method presented combines recent advancements in multi-field visualization as well as topological data analysis for visualization, namely merge trees. In the following, we will briefly put our work into the context of existing contributions.
\subsection*{Attribute space interactions for multi-field visualization}
Coordinated multiple linked views are often used to cope with multi-fields or other complex data, see e.g. the state-of-the-art report by Roberts et al.~\cite{Roberts2007}. In most of this work, representations in attribute space play an as important role as spatial representations. In this context, multi-dimensional transfer function~(TF) design plays an important role to generate spatial representations~\cite{Ljung2016}.
To reduce the complexity of the task a common approach is to use attribute space clustering or segmentation.
Wang et al.~\cite{Wang2012} propose to segment a 2D density plot in attribute space by facilitating a Morse decomposition to automatically generate a TF.
With a similar goal, Cai et al.~\cite{Cai2017} suggest a two-level approach, starting with a topology-preserving dimensionality reduction step followed by a clustering step. Another cluster-hierarchy-based method used for interactive transfer function generation was proposed by Dobrev et al. ~\cite{Dobrev2011}. Here a cluster tree visualization in combination with parallel coordinates serves as an interactive interface.

Given the multidimensional nature of tensors, visualization methods for multi-variate data have also been developed in context with tensor field visualization.
Kindlmann et al.~\cite{kindlmanvolume} introduced direct volume rendering (DVR) for tensor fields using a barycentric shape space for opacity assignment.
Jankowai et al.~\cite{Jankowai2020a} present an interface using glyph widgets to design a transfer function for rendering tensor features. They augment the volume rendering with a texture for directional information. The widgets employ characteristic glyph representatives for intuitive navigation through the attribute space.
Refer to~\cite{Kratz2013,tensorstar} for respective overview articles.

%
\subsection*{Topology guided visualisation}
Concepts from topological data analysis can be found in many visualization applications for scalar fields. Especially structures like the contour tree or merge tree are frequently used to guide visualizations in an increasing number of applications. In an early paper, van Kreveld et al.~\cite{vanKreveld1997a} augment a contour tree with seed sets for fast iso-contour computation. Weber et al.~\cite{weber2007a} present an approach for volume rendering of topologically segmented scalar fields, assigning a different transfer function to each segment. Takahashi et al.~\cite{Takahashi2004} follow a similar goal proposing automatic transfer functions accentuating topological changes in scalar fields. Takeshima et al.~\cite{Takeshima2005a} have extended this work by introducing a set of topological attributes that serve as auxiliary variables for the design of multi-dimensional transfer functions.
Especially successful are methods that integrate results from the topological analysis into interactive frameworks.
Bremer et al.~\cite{Bremer2011} offer a linked-view interface for efficient analysis of burning cells from turbulent combustion simulations. They augment the computed merge tree with global and local statistical information, supporting extensive feature extraction and analysis. Bock et al.~\cite{Bock2018} also used a combination of merge tree analysis and an interactive user interface for efficient segmentation of micro-CT scan data of fishes. Besides the merge tree, other topological structures have also been used in visual frameworks. An example is the work by Shivashankar et al.~\cite{Felix} who present a queryable hierarchy of Morse-Smale complexes that allows astronomers to examine filamentary structures of the cosmic web on different scales. A survey of topology-based methods in visualization can be found in the report by Heine et al.~\cite{topologystar}.
\subsection*{Extension of iso-surfaces and topological concepts to multi-fields}
A core feature of the presented method is the rendering of relevant iso-surfaces for multi-fields. Multi-field iso-surfaces have initially become accessible through the work of Carr et al. \cite{fibersurfaces}. They have introduced an extension of iso-surfaces to bi-variate data using sets of fibers. Fibers can be understood as the bi-variate equivalent to iso-lines which allows for the generation of fiber surfaces based on polygons defined in attribute space.
Different extraction and rendering methods for fiber surfaces have since been presented.
Wu et al. \cite{Wu2017a} offers a system for interactive exploration of bi-variate data through real-time pixel-perfect fiber surface rendering. This is achieved through intersection tests in range space to calculate fiber surfaces on the fly. Klacansky et al. \cite{Klacansky2017a} implemented a topology-agnostic and exact calculation of fiber surfaces with a speedup of up to two orders of magnitude compared to the initial article using a range of acceleration methods.
Fiber surfaces have then been generalized to general multi-variate data by Raith and Blecha et al. \cite{raithsalamislice1,raithsalamislice2}. In their framework, user-defined geometries in attribute space are called interactors. Intersections of input cells with this geometry in attribute space define the set of points in the spatial domain that form the iso-surface.
The latest addition to multi-field iso-surface visualisation has been feature level-sets (FLS) \cite{JankowaiHotz2019}. They are an approach to generalising iso-surfaces to multi-variate data via distance field computation in attribute space. For a given set of points in attribute space, the distance towards a user-defined geometry or range is calculated. The geometry or range is called a trait and defines the parameters of interest. Once the distance field computation is finalised, the resulting field is pulled back into domain space and can be used to render iso-surfaces.
FLS have been applied by Nguyen et al. \cite{Nguyen2021a} and Athawale et al. \cite{Athawale2021a} to separate and visualise large- and small-scale structures in Taylor-Couette flow simulation data and to examine the correlation between velocity magnitude and temperature attributes in numerical simulations of idealised solar farms, respectively.

\section{Data, traits, and merge trees}

\begin{figure*}
    \centering
    \includegraphics[width=\linewidth]{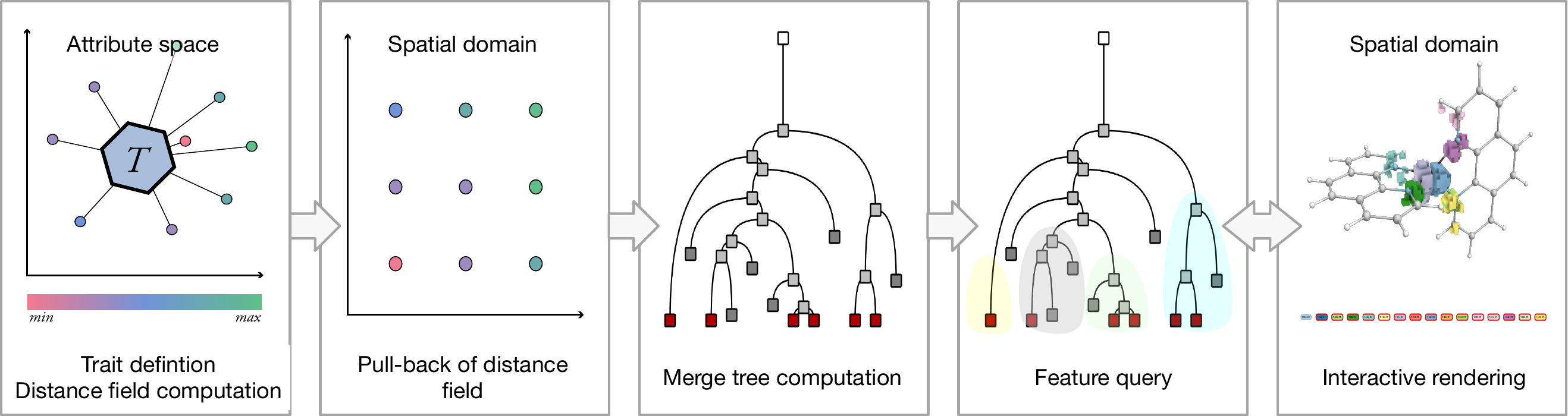}
    \caption{Pipeline schematic. The user-defined trait corresponds to the parameter settings that are of interest. For every vertex, the distance to this trait is calculated in attribute space, whereafter the distance value is assigned to the vertex in the domain (pull-back). This concludes the feature level-set computation. The resulting distance field serves as input for the computation of a merge tree. The tree undergoes simplification and can then be queried in several ways, see \sref{sec:query_methods}. Finally, the user can interact with the resulting domain segmentation via a legend or a slice through the data.}
    \label{fig:pipline}
\end{figure*}
\label{sec:background}
The proposed method is based on three main concepts. Multi-variate data, feature level-sets and traits, and merge trees. A summary of these concepts is given below. For a more extensive definition, we refer the reader to \cite{JankowaiHotz2019} and \cite{Carr2003a}, respectively.
\subsection{Multi-variate data}
We assume a multi-variate data as input, given as a set of continuous fields $F_1, F_2, \cdots, F_m: \Xspace \to \Rspace$ defined on the \emph{data domain} $\Xspace$. In the case of tensor fields, these functions are the components of the tensor field. We then assemble an \emph{attribute space} $\Aspace \subset \Rspace^n$ by combining a set of (selected) field values and possibly some derived quantities. Such a data set is then transformed and summarised by a multi-variate mapping, $f: \Xspace \to \Aspace$, where $n$ is the number of selected field values and their derived quantities.
The attribute space $\Aspace$ is equipped with a metric (e.g., a Euclidean metric), denoted as $d_{\Aspace}$.
\subsection{Feature level sets}
We define a \emph{trait} $T$ as a subset in the attribute space $T \subset \Aspace$.
$T$ may. e.g., be a convex polygon, a point, a collection of points, or a line segment. A \emph{feature} in the data domain is the pre-image of a trait from the attribute space, $f^{-1}(T) = \{x \in \Xspace \mid f(x) \in T \subset \Aspace\}$.
For an arbitrary trait in an attribute space of arbitrary dimension, its corresponding feature in the data domain may be empty.
Therefore, for a fixed trait $T \subset \Aspace$, we define the \emph{trait distance field}, $d_T: \Aspace \to \Rspace$, where $d_T(a) = \min_{t \in T} d_{\Aspace}(a, t)$.
The \emph{trait-induced distance field} (or feature distance field) is a scalar function defined as $h_T = d_T \circ f: \Xspace \to \Rspace$.
Finally, the \emph{trait-induced level sets} are the level sets of $h_T$,
$h_T^{-1}(c) = \{x \in \Xspace \mid h_T(x) = c\}$.
\subsection{Merge trees}
Let  $g: \Xspace \to \Rspace$ be a continuous scalar field. 
For the computational purpose, assume $g$ is defined on a simply connected compact simplicial complex $\Xspace$ and is linearly interpolated on the interiors of its simplices.
Two points $x, y \in \Xspace$ are considered \emph{equivalent}, denoted by $x \sim y$, if $g(x) = g(y)$ and $x$ and $y$ are a part of the same connected component of the sub-level set, i.e. $g^{-1}(-\infty, g(x)] = g^{-1}(-\infty, f(y)]$. The quotient space $\Xspace/\sim$ is called a \emph{merge tree} of $g$. The merge tree records birth, death, and merge events of sub-level set components during a sweep of $g$ from $-\infty$ to $\infty$.
Typically, merge trees are computed using algorithms based on the work by Carr et al.~\cite{Carr2003a}. It is based on a \emph{sub-level set filtration} of $g$, which is observing changes in a sequence of nested sub-level sets connected by inclusions.

\subsection{Merge tree simplification}
\label{sec:simplification}
The original tree may contain many leaves coming from noisy data, imprecision, or irrelevant structures.
Two common metrics, that we consider in our pipeline, for determining which leaves and arcs should be simplified (merged into their parent arc) are persistence and hypervolume.
\newcommand\metricfigurewidth{0.43\linewidth}
\begin{figure}
    \centering
    \subfloat[\label{subfig:metric_persistence}]{
        \includegraphics[width=\metricfigurewidth]{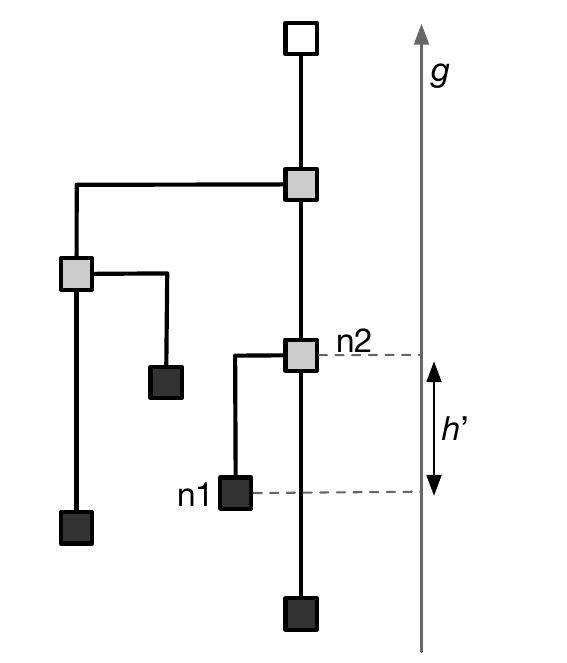}}
    \subfloat[\label{subfig:metric_hypervolume}]{
        \includegraphics[width=\metricfigurewidth]{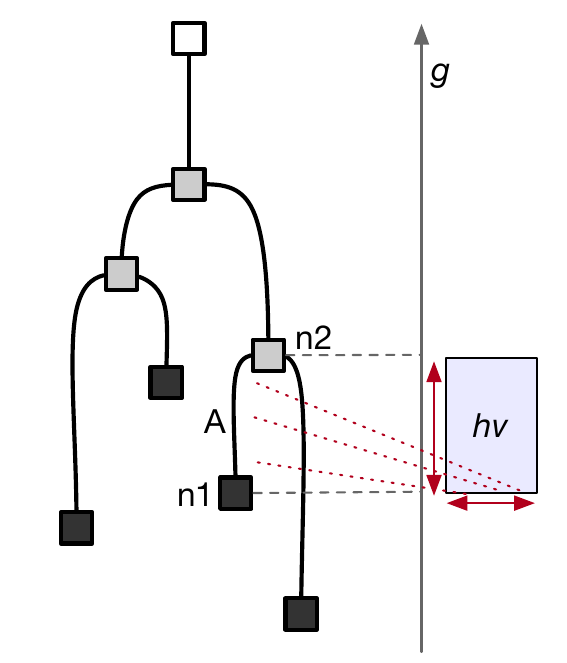}}
    \caption{Simplification metrics: \protect\subref{subfig:metric_persistence} Persistence. Persistence $h'= g(n_2)-g(n_1)$ where $g(x)$ is the scalar function over the domain. Where $n_1$ and $n_2$ are two paired critical points. \protect\subref{subfig:metric_hypervolume} Hypervolume. For any arc $A$, the volume of all associated vertices is accumulated and multiplied by the height of the corresponding arc. Hence, hypervolume is a metric that takes the spatial embedding of the data into account while persistence is only related to the data range.}
    \label{fig:metric}
\end{figure}
\paragraph{Persistence}
In context with merge trees, persistence arises from a \emph{sub-level set filtration} generating a pairing of critical points $(x_i, y_i)$. During filtration, a feature is generated in one critical point and disappears in the other. The interval spanned by the function values of the critical points $[g(x_i) = g(y_i)]$ represents the feature lifetime interval. To each pair of critical points, one can then assign a \emph{ persistence value} which is the difference of the scalar values in the two critical points $g(x_i)-g(y_i)$. It gives some notion  feature stability~\cite{Cohen-SteinerEdelsbrunnerHarer2010}. This critical point pairing can be used for a controlled simplification of the data removing features ordered by their persistence value.
%
A \emph{branch decomposition tree} derived from a merge tree is a hierarchical representation of these pairs of critical points~\cite{Pascucci2005a}, see \fref{fig:query}b.

The geometric interpretation of a low persistence feature in a two-dimensional example would be a shallow valley on a height-field map. Likewise, a high persistence value equates to a deep intrusion in a field. Low persistence values often occur in noise where small differences in function values create irrelevant extremal points. To retain the relevant features while eliminating irrelevant ones, low-persistent branches, leaves, and their incident arcs are merged into their parent branch.
As shown in \fref{subfig:metric_persistence}, persistence directly correlates to tree height.
\paragraph{Hypervolume}
Hypervolume \cite{Bock2018} is a metric that takes local geometric measures of the underlying data domain into account. As shown in \fref{subfig:metric_hypervolume}, hypervolume is the product of arc height and the volume contained by the region corresponding to the arc. In practice, this is the number of voxels contained in a segment multiplied by the difference in function values at the minimum and the saddle connected by the arc. Unlike persistence, the hypervolume metric requires additional information about vertex-segment association for computation.
%

%

%

\section{Trait-induced merge trees}
\label{sec:construction}
Feature level sets (FLS) enable the user to extract features as iso-surfaces according to a certain feature distance interest. They do not, however, offer a way to perform any sort of selection or filtering which may result in a cluttered rendering or the inclusion of noise in the rendering.
Combining FLS with scalar field topology, merge trees in particular, adds the option to select and filter surfaces based on some metric, see \sref{sec:simplification}. The use of merge trees comes naturally as their leaves correlate to those data points in the data which are closest to the defined trait.
For a fixed trait $T \subset \Aspace$, given its trait-induced distance field $h = h_T: \Xspace \to \Rspace$, we can define a \emph{trait-induced merge tree} by tracking how the level set components $h^{-1}(c)$ merge as we vary the distance parameter $c$.
Similarly, we obtain a \emph{trait-induced merge tree} by tracking the evolution of the sub-level sets $h^{-1}(\infty, c]$. Such a trait-induced merge tree may be used directly (or indirectly) as an interface to guide the exploration of multivariate data.
\newcommand\queryfigurewidth{\linewidth}
\begin{figure}
    \centering
        \includegraphics[width=\queryfigurewidth]{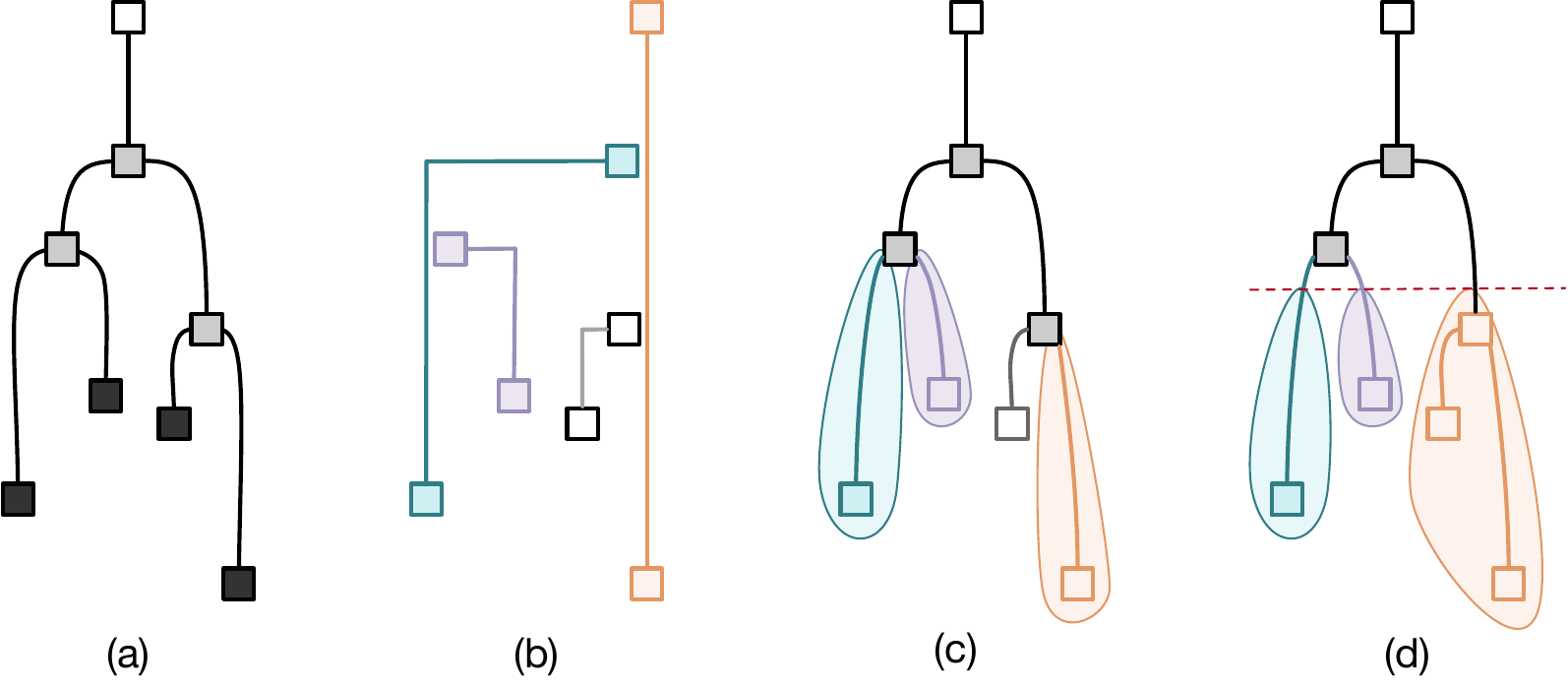}
    \caption{Query methods. (a) shows the original tree. The user may choose between segmentation of the tree based on (b) branch decomposition, (c) leaf nodes, (d) sub-trees. Every query method has its merits and drawbacks, see \sref{sec:query_methods}.}
    \label{fig:query}
\end{figure}

\subsection{Merge tree-based features}
\label{sec:query_methods}
Data sets as well as research questions are different in nature.
Therefore, we provide a number of different simplification and query methods for the merge tree, each of which emphasizes different aspects of the data.
\paragraph{Branch decomposition} The branch decomposition is a common representation for merge trees (see \fref{fig:query}b). It allows for hierarchical simplification and querying. When using this representation as a basis for segmentation, the user specifies a threshold for the simplification (persistence or hypervolume) and the method returns a segmentation of the domain according to the simplification of the tree at that threshold. This method always contains one branch which connects the global minimum to the maximum. In terms of visualization, this branch is often problematic since its vertices usually enclose all others. When rendering the segment of this branch, all other segments as well as the global minimum may not be visible.
\paragraph{Extremal points and their incident arcs} This method extracts the leaf nodes and their neighboring vertices and segments the domain accordingly (see \fref{fig:query}c). Here too, the user first specifies a simplification threshold. Unlike in the first method, it is now the merge tree itself that is simplified and then queried. This method has the advantage of highlighting every minimum separately which gives a detailed overview of the spatial distribution of minima.
\paragraph{Sub-trees} Sub-trees are extracted by first simplifying the tree as above and then cutting it at a user-specified level (see \fref{fig:query}d). The segments are then given by the vertices whose function value is below the threshold and which are contained within the branches that are directly affected by the cut. The result is something similar to contour forests where each segment in the domain is specified by a sub-tree originating from the cut downward.
\section{Interaction and rendering}
\label{sec:interaction}
The method has been implemented in an interactive visualization framework using Inviwo~\cite{Jonsson2020b} providing a large variety of rendering options. Besides specifying the rendering options the main interaction possibilities concern the trait specification and the feature selection.

\paragraph{Trait specification}
For the trait specification, we provide similar options as the ones introduced in the original paper about feature level sets~\cite{JankowaiHotz2019}. In the current implementation, we support point traits which consist of one or a few points in attribute space; cubical traits, defined by intervals in all attribute space dimensions; or a few further explicit geometries. As the main interaction panel, we use a parallel coordinates plot where one line crossing all parallel axis corresponds to a point trait and a set of intervals for the cubical traits. For tensor field, we additionally support picking of tensors in a glyph rendering of the data set.

\paragraph{Feature selection}
For the feature, we provide two ways of interacting with the segmentation of the domain. The first interface consists of a legend positioned below the 3D rendering. The legend contains a button for every segment in the data. Clicking on such a button will toggle the voxel-wise rendering of that segment. Active buttons are highlighted with a red boundary. For navigation, the colors of the buttons and the rendered segments correspond to each other. Additionally, the buttons show the value of the minimum of the segment, which is a measure of the distance to the trait. This way, the user can select segments based on their distance to the trait.

%


\section{Results}
\label{sec:results}

We demonstrate the powerful utility and generality of our proposed framework across a range of scalar, vector, and tensor field data sets.

\subsection{Two-point load}
\label{sec:tensor}
\newcommand\tplfigurewidth{0.48\linewidth}
\begin{figure}[ht]
    \centering
    \subfloat[\label{fig:tpl-illustration}]{
        \includegraphics[width=\tplfigurewidth]{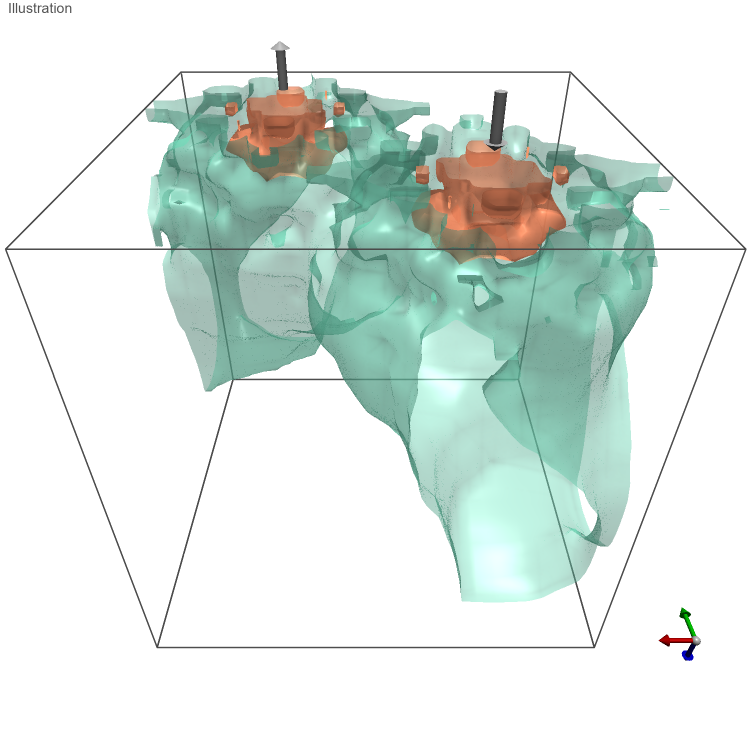}
    }
    \hfill
    \subfloat[\label{fig:tpl-zero-eigenvalues}]{
        \includegraphics[width=\tplfigurewidth]{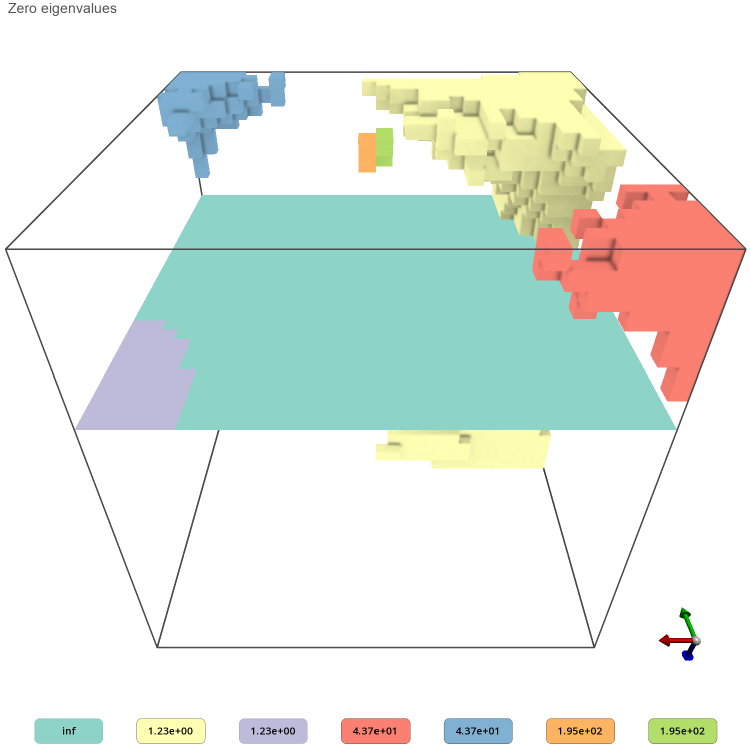}
    }

    \subfloat[\label{fig:tpl-spherical}]{
        \includegraphics[width=\tplfigurewidth]{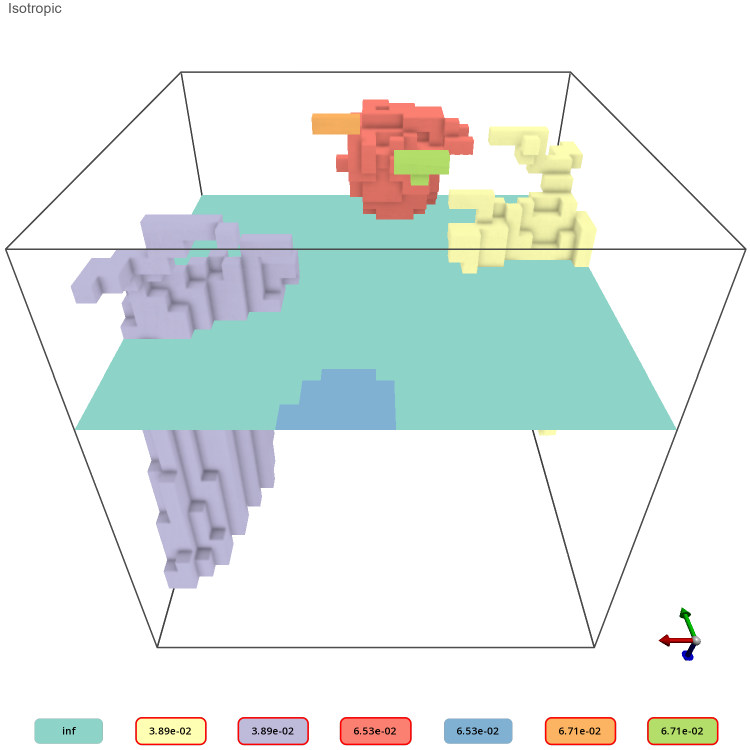}
    }
    \hfill
    \subfloat[\label{fig:tpl-planar-lambda}]{
        \includegraphics[width=\tplfigurewidth]{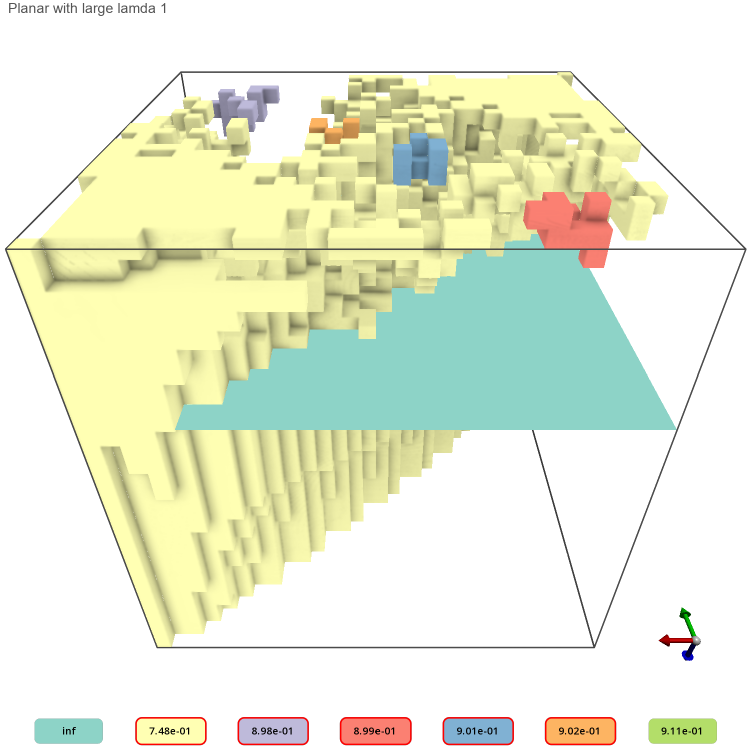}
    }

    \caption{
        \protect\subref{fig:tpl-illustration} Illustration of the two-point-load data set. The arrows indicate the forces applied to the block of metal. The iso-surfaces display the anisotropy that occurs in the material.
        \protect\subref{fig:tpl-zero-eigenvalues} shows segmentation of distance field for regions where the eigenvalues are close to zero.
        \protect\subref{fig:tpl-spherical} shows segmentation of distance field for high spherical anisotropy, i.e. isotropic behaviour.
        \protect\subref{fig:tpl-planar-lambda} shows segmentation of distance field for planar behaviour coupled with a high principal eigenvalue.
        The active segment buttons are highlighted with a red boundary.
    }
    \label{fig:tpl}
\end{figure}
%
Since FLS was originally developed with tensor fields in mind, our first example is based on tensor fields. Using an individual tensor as a trait, FLS have introduced a notion of ``tensor iso-surfaces''. In combination with trait-induced merge trees, this opens an entirely new concept of ``tensor field topology''.
To verify the proposed method, the first case study involves a well-known numerical material simulation of stresses inside a solid block.
Onto the top of this block, two forces are applied, one pulling and one pushing, as illustrated in \fref{fig:tpl-illustration}.
Subsequently, this data set will be referred to as `two-point-load'.
The output of the simulation is a stress tensor-field containing symmetric tensors which have six independent degrees of freedom.
As described in \fref{fig:pipline}, the first step is to define traits and calculate the distance field.
The expected material stress is highly anisotropic at the points of impact, unaffected in areas furthest from the impact points and planar along the midsection between the impact points.
In order to express these states, we will use the following terminology: Eigenvalues are referred to as principal stresses $\lambda_i$. They are ordered such that $\lambda_1\ge\lambda_2\ge\lambda_3$, named major, intermediate, and minor principal stress.
Linear, planar, and spherical anisotropies are given by $c_l=\frac{\lambda_1-\lambda_2}{\lambda}$, $c_p=\frac{2(\lambda_2-\lambda_3)}{\lambda}$, and $c_s=\frac{3\lambda_3}{\lambda}$ respectively with $\lambda=\lambda_1+\lambda_2+\lambda_3$ \cite{Westin1997}.
One frequently used invariant is the maximum shear stress that is defined as $\lambda_1-\lambda_3$. \fref{fig:tpl-illustration} shows two level-sets of this invariant.
The traits were defined to match these criteria to verify the resulting regions.
For the the first trait, all three principal stresses were set to zero. \fref{fig:tpl-zero-eigenvalues} shows regions that are close to this behaviour. The result is in line with our expectations since the highlighted regions are those farthest away from the impact points.
\fref{fig:tpl-spherical} shows regions with isotropic behaviour. The corresponding trait was set to high spherical anisotropy ($c_s$) and low linear and planar anisotropies ($c_l$ and $c_p$). Comparing it to \fref{fig:tpl-zero-eigenvalues}, we can see that the regions are neighbouring which makes sense. Regions that are entirely unaffected by the forces will have a small band of rather isotropic behaviour around them before a more distinctive stress distribution emerges.
Lastly, the third trait was set to highly planar behaviour ($c_p$) coupled with a high major principal stress value ($\lambda_1$). \fref{fig:tpl-planar-lambda} depicts the resulting regions. As expected, the middle section corresponds to this trait.

\subsection{Extraction of acceptor and donor regions in molecular electronic transitions}
\label{sec:chemistry}
\newcommand\molfigurewidth{0.22\linewidth}
\begin{figure*}
    \centering
    \subfloat[\label{subfig:mol-symmetric-donor-csp}]{
        \includegraphics[width=\molfigurewidth]{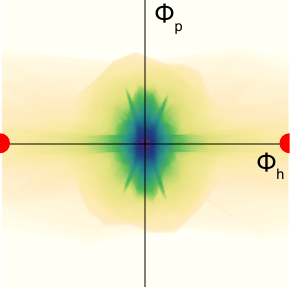}
    }
    \hfill
    \subfloat[\label{subfig:mol-symmetric-donor-df}]{
        \includegraphics[width=\molfigurewidth]{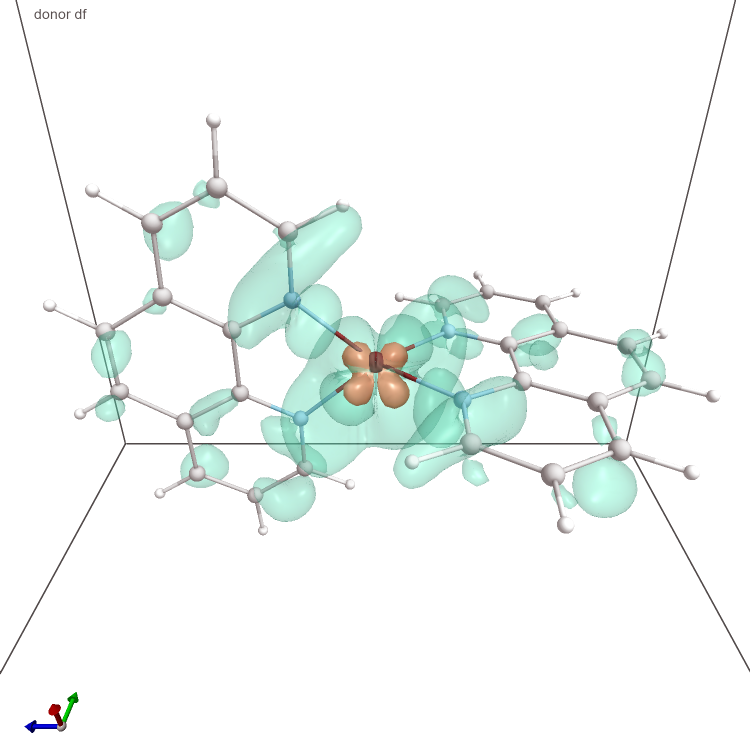}
    }
    \hfill
    \subfloat[\label{subfig:mol-symmetric-donor}]{
        \includegraphics[width=\molfigurewidth]{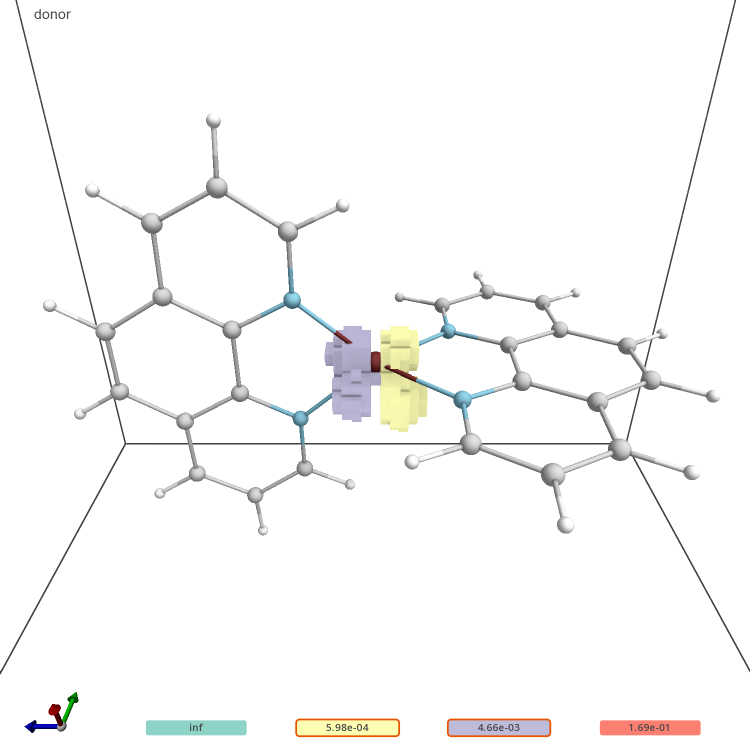}
    }

    \subfloat[\label{subfig:mol-symmetric-acceptor-csp}]{
        \includegraphics[width=\molfigurewidth]{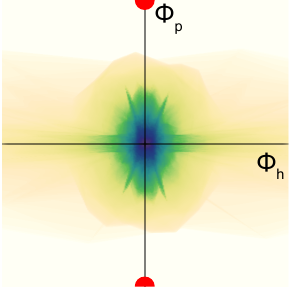}
    }
    \hfill
    \subfloat[\label{subfig:mol-symmetric-acceptor-df}]{
        \includegraphics[width=\molfigurewidth]{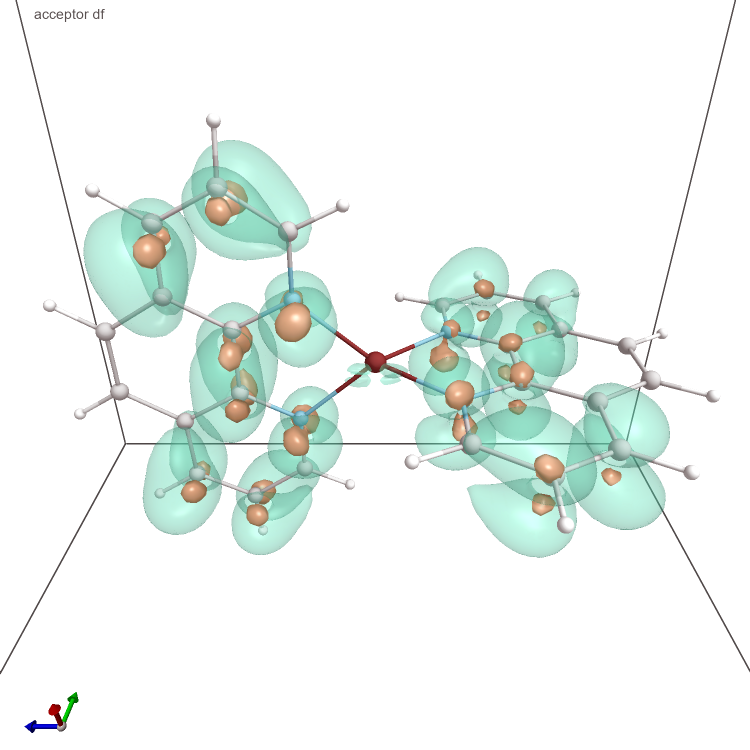}
    }
    \hfill
    \subfloat[\label{subfig:mol-symmetric-acceptor}]{
        \includegraphics[width=\molfigurewidth]{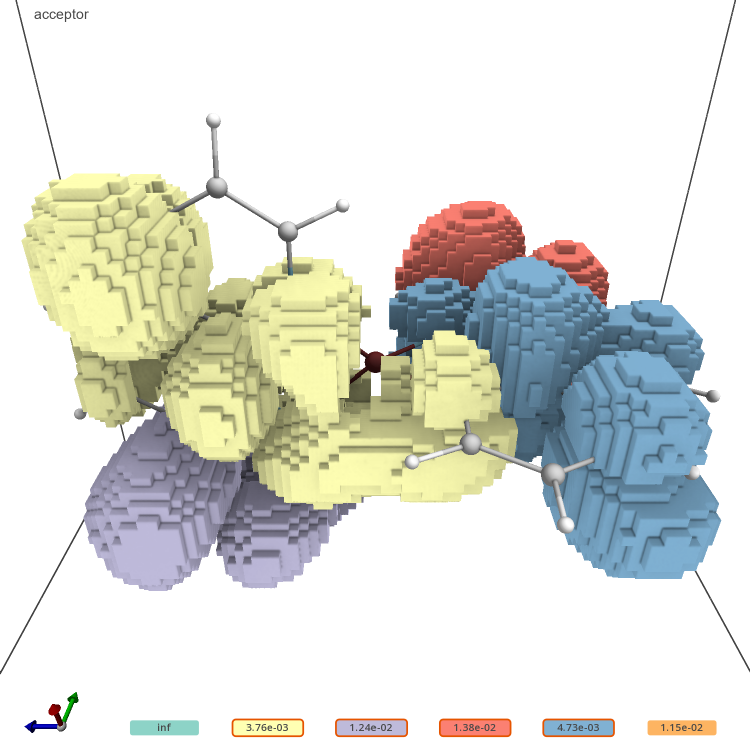}
    }

    \caption{Molecular electronic transition data set (copper complex) with symmetric ligands.
        \protect\subref{subfig:mol-symmetric-donor-csp} and
        \protect\subref{subfig:mol-symmetric-acceptor-csp} CSP plots and the donor and acceptor trait, respectively (red dots).
        \protect\subref{subfig:mol-symmetric-donor-df} donor and acceptor distance field obtained from the FLS computation.
        \protect\subref{subfig:mol-symmetric-donor} shows donor regions while \protect\subref{subfig:mol-symmetric-acceptor} shows acceptor regions.
        The merge tree has been simplified using the hypervolume metric. The extracted regions correspond to the $n$ lowest leaves in the tree.}
    \label{fig:mol-symmetric}
\end{figure*}
\begin{figure*}
    \centering
    \subfloat[\label{subfig:mol-asymmetric-donor-csp}]{
        \includegraphics[width=\molfigurewidth]{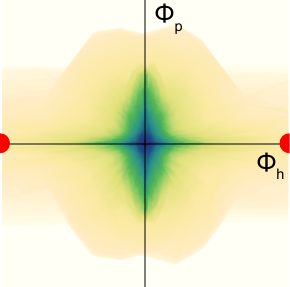}
    }
    \hfill
    \subfloat[\label{subfig:mol-asymmetric-donor-df}]{
        \includegraphics[width=\molfigurewidth]{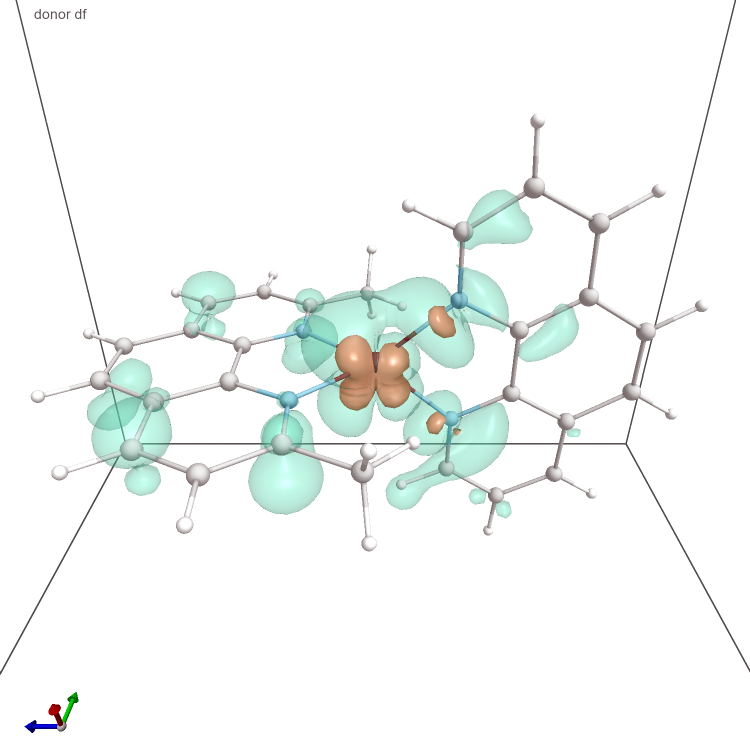}
    }
    \hfill
    \subfloat[\label{subfig:mol-asymmetric-donor}]{
        \includegraphics[width=\molfigurewidth]{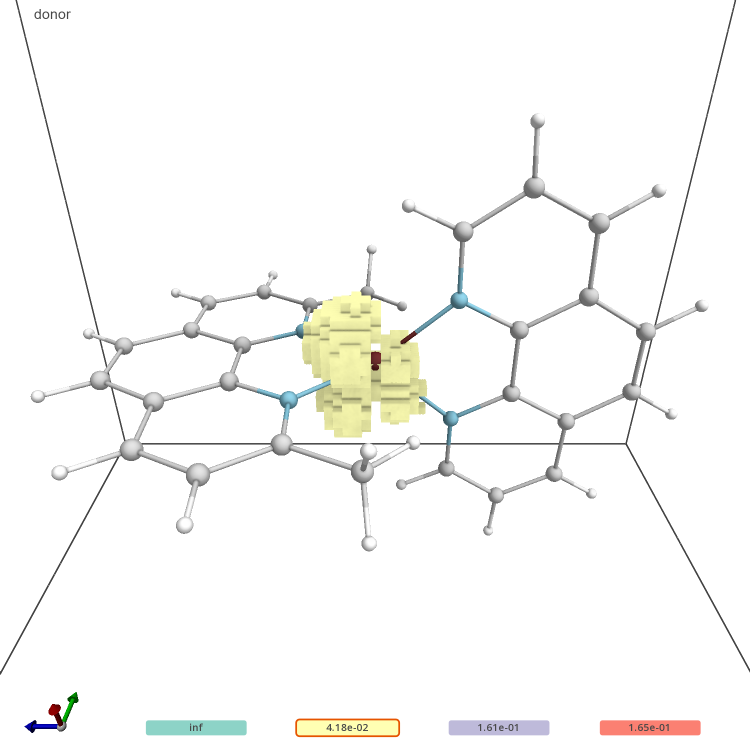}
    }

    \subfloat[\label{subfig:mol-asymmetric-acceptor-csp}]{
        \includegraphics[width=\molfigurewidth]{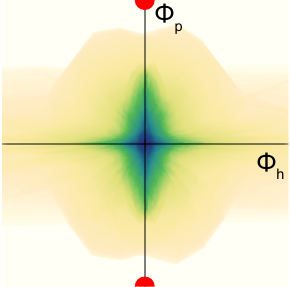}
    }
    \hfill
    \subfloat[\label{subfig:mol-asymmetric-acceptor-df}]{
        \includegraphics[width=\molfigurewidth]{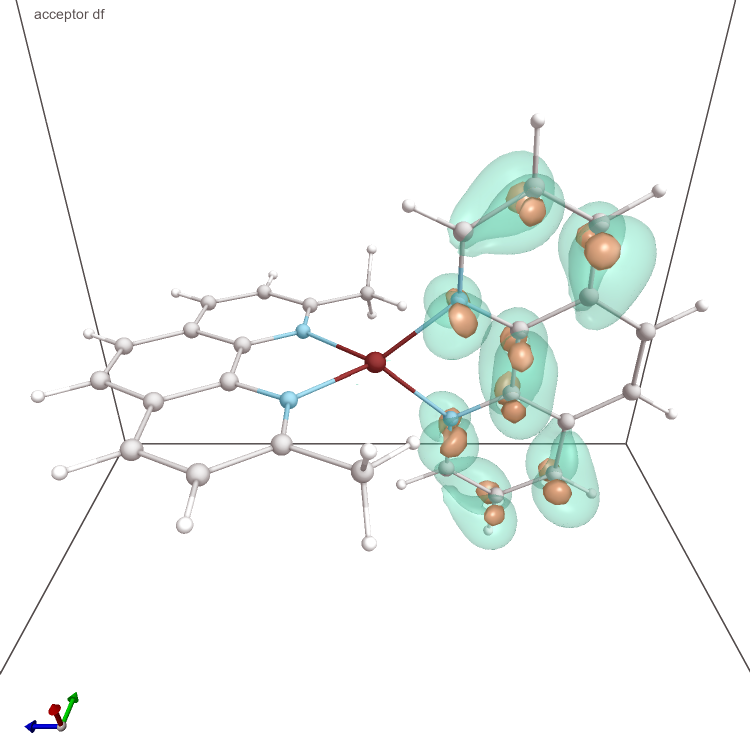}
    }
    \hfill
    \subfloat[\label{subfig:mol-asymmetric-acceptor}]{
        \includegraphics[width=\molfigurewidth]{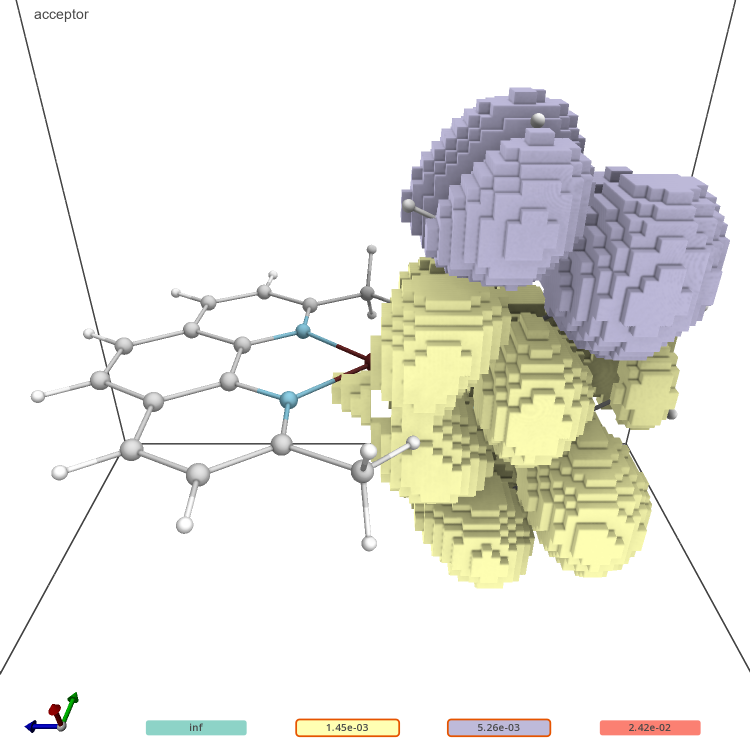}
    }

    \caption{Molecular electronic transition data set (copper complex) with asymmetric ligands.
        \protect\subref{subfig:mol-symmetric-donor-csp} and
        \protect\subref{subfig:mol-symmetric-acceptor-csp} CSP plots and the donor and acceptor trait, respectively (red dots).
        \protect\subref{subfig:mol-symmetric-donor-df} donor and acceptor distance field obtained from the FLS computation.
        \protect\subref{subfig:mol-symmetric-donor} shows donor regions while \protect\subref{subfig:mol-symmetric-acceptor} shows acceptor regions.
        The merge tree has been simplified using the hypervolume metric. The extracted regions correspond to the $n$ lowest leaves in the tree.}
    \label{fig:mol-asymmetric}
\end{figure*}
In this case study, we analyse the molecular electronic transitions using feature level-sets and trait-induced merge trees. The electronic structure of a molecule undergoes a change on interaction with light. This change can be represented concisely by two scalar fields, $\Phi_h$ and $\Phi_p$, denoting the spatial distribution of the electron before and after absorption of photon during the electronic transition~\cite{Masood2021Transitions}. Chemists are interested in studying how the localisation of the electronic distribution changes during the transition and how different molecular configurations affect the transitions. In particular, it is crucial to identify which parts of the molecule lose and gain charge, and therefore, act as donor and acceptor regions within the molecule, respectively. Recently, Sharma~\etal\cite{Sharma2021CSPSegmentation, Sharma2023CSPOperators} proposed considering the two scalar fields corresponding to an electronic transition as a single multi-field and applied bi-variate analysis on the resulting field. They suggested examining the patterns in the continuous scatter plots~\cite{Bachthaler2008CSP} of the bi-variate field corresponding to the complete molecule or sub-regions within the molecule can help reveal the donor and acceptor behaviour.

Here, we consider electronic transitions in two copper complexes with slightly different configurations. One of these complexes has the same two molecular groups around the central copper atom, a case of \emph{symmetric ligands}. The other complex has two different molecular groups around the copper atom, a case of \emph{asymmetric ligands}. The task of interest to the chemists is to identify the donor and acceptor regions within these two complexes and compare if and how these two configurations differ. To accomplish this, we first convert the notion of a donor and acceptor into well-defined traits. A donor is a subset of the molecule where there is more concentration of electronic density before the transition compared to the state of the electronic distribution after transition. That is, in general, a donor region can be characterised by the set of points satisfying the condition $|\Phi_h| > |\Phi_p|$. However, in the case of an ideal donor behaviour, we expect $|\Phi_p|$ to be zero while $|\Phi_h|$ is simultaneously very high. Therefore, we define the donor trait as a set of two points with coordinates $(\max|\Phi_h|, 0)$ and $(-\max|\Phi_h|, 0)$ in the bi-variate range space spanned by $\Phi_h \times \Phi_p$. This point set trait is indicated by two red disks in \fref{subfig:mol-symmetric-donor-csp} and \fref{subfig:mol-asymmetric-donor-csp}. Following this idea, we define the acceptor by a trait consisting of two points, $(0, \max|\Phi_p|)$ and $(0, -\max|\Phi_p|)$, see \fref{subfig:mol-symmetric-acceptor-csp} and \fref{subfig:mol-asymmetric-acceptor-csp}.

With these point traits for donor and acceptor regions, we extract the feature level-sets for both molecules. In the case of symmetric ligands, the level-set corresponding to the donor trait concentrates around the central copper atom, as can be observed in \fref{subfig:mol-symmetric-donor-df}. This is as expected by the chemists, since the copper atom is known to act as a strong donor in these complexes. The feature level-sets corresponding to the acceptor trait are, however, distributed over the surrounding two molecular groups, see \fref{subfig:mol-symmetric-acceptor-df}. This behaviour is also as expected because there is no reason for the electron to prefer one molecular group over the other, both groups being the same. These feature level-sets extracted for the donor and acceptor traits were very appreciated by our collaborating partner who shared this data with us and is an expert in this domain. Since selecting the right distance threshold for feature extraction can be tricky and involves interactive visual exploration of the data, we further applied the analysis based on trait-induced merge trees to this transition. We queried the regions corresponding to the leaf nodes with lowest values in the merge tree of the donor trait and automatically found the region concentrated around the copper atom, as shown in \fref{subfig:mol-symmetric-donor}. Similarly, for the acceptor trait, the regions are distributed over the two groups around the copper atom as shown in \fref{subfig:mol-symmetric-acceptor}. Interestingly, these regions corresponding to leaves with lowest values closely capture the two molecular groups as separate regions in the merge tree segmentation. This automatic subdivision of the acceptor region into sub-regions matching the chemical subgroups in the molecule was also appreciated by our collaborator.

Next, we repeat the same analysis on the second copper complex with asymmetric ligands. As was the case previously, the donor region is still concentrated on the central copper atom, as shown in \fref{subfig:mol-asymmetric-donor}. However, the acceptor region is now concentrated on only one of the two surrounding molecular groups as can be seen in \fref{subfig:mol-asymmetric-acceptor}. This suggests, in the case of asymmetric ligands, electronic charge transfer can occur with preference for one molecular group over the other. As mentioned before, such change in behaviour of charge transfer in transition due to change in molecular configuration is of particular interest to the chemists, and this analysis is possible with our proposed techniques. Lastly, we want to point out that the continuous scatter plots for both the complexes look quite similar, as seen in \fref{subfig:mol-symmetric-donor-csp} and \fref{subfig:mol-asymmetric-donor-csp}. Therefore, it is difficult to distinguish between the character of these two transitions just based on these plots, there is a need to explore the spatial domain as well which is facilitated by the feature level-sets and the domain segmentation based on trait-induced merge trees.
\subsection{Vortex re-connection}
\label{sec:flow}
\newcommand\flowfigurewidth{0.98\linewidth}
\begin{figure}[t]
    \centering
    \subfloat[\label{fig:flow-df}]{
        \includegraphics[width=\flowfigurewidth]{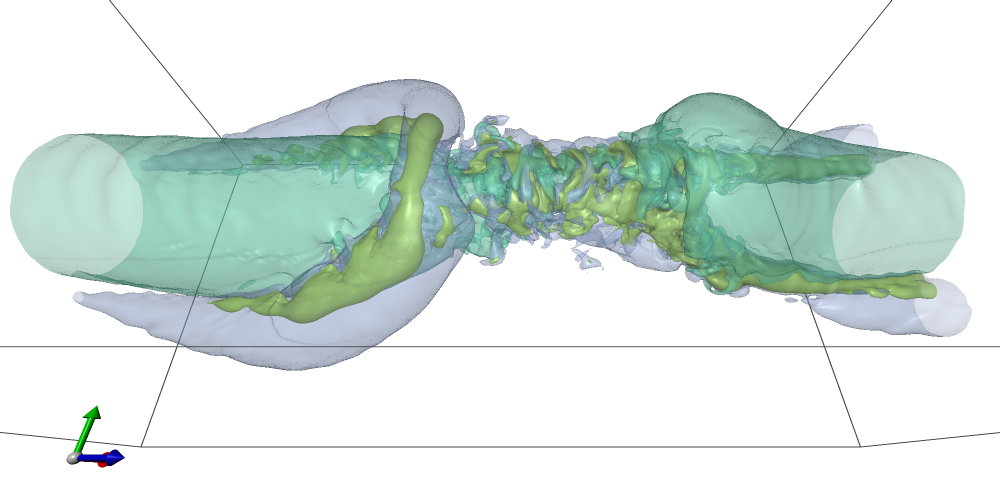}
    }

    \subfloat[\label{fig:flow-tf}]{
        \includegraphics[width=\flowfigurewidth]{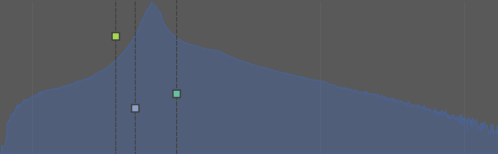}
    }

    \subfloat[\label{fig:flow-seg}]{
        \includegraphics[width=\flowfigurewidth]{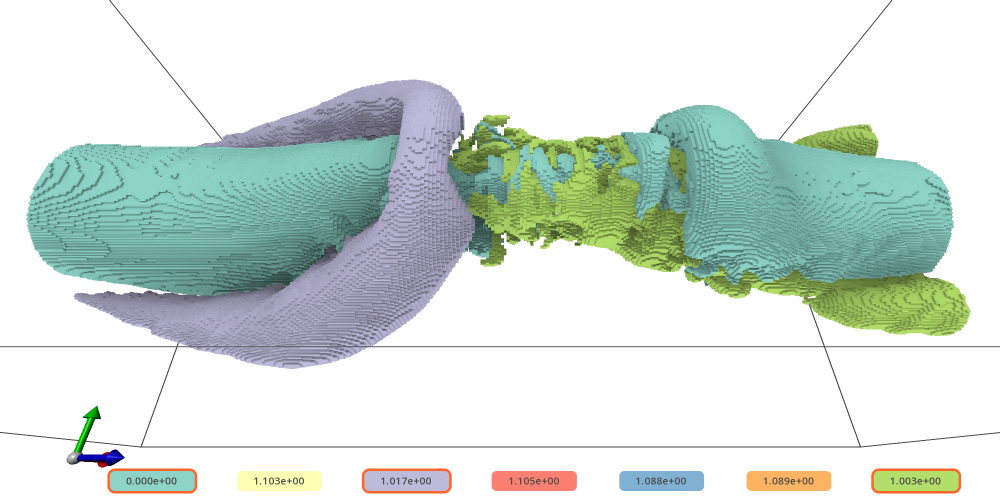}
    }
    \caption{Vortex re-connection simulation.
        \protect\subref{fig:flow-df} shows volume rendering of the distance field.
        \protect\subref{fig:flow-tf} shows histogram and iso-surface placement. The histogram of the distance field is given by the blue graph in the background. Distance from the trait is mapped to the x-axis while opacity is mapped to the y-axis.
        \protect\subref{fig:flow-seg} shows segmentation using branch decomposition.}
    \label{fig:flow}
\end{figure}

Vortices are regions of rotational fluid flow around a core line defined by an arbitrary curve in space. They play a fundamental role in the study of fluid dynamics. There are numerous methods for the identification and extraction of vortex structures and their core lines~\cite{Gunther2018Vortex, Kasten2011Acceleration}. However, the dynamics of vortices -- how they evolve and interact with each other -- is less understood, and therefore, this topic is still under active research~\cite{Bujack2020FlowSTAR, Kasten2012VortexMerge}. The data we examine as part of this case study concerns the interaction of two parallel counter-rotating vortices. Over time, these two vortices come closer and go through a \emph{re-connection} event. This phenomenon can be observed, for example, in the vortices shed by the wing tips of an airplane. These vortices can sometimes be seen in the sky as condensation trails. The trails close to the airplane are initially parallel; then followed by the re-connection event, they form closed loop-like structures; before they dissipate completely in the air. Our collaborators performed a numerical simulation of this phenomenon. One of their main goals for conducting this simulation is to understand the re-connection event. It is known that after re-connection of two counter-rotating vortices, a \emph{horseshoe} like structure emerges at the re-connection point which grows bigger in size with time and finally results in the formation of disconnected loops. Thus, the identification and analysis of this horseshoe structure is a key task in the analysis of this particular simulation data.

Unlike vortices and their cores, there is no clear definition available that we can directly apply to automatically extract these horseshoe structures. We therefore explored whether feature level sets and the trait-induced merge tree are useful for this task. Based on input from the domain experts who conducted this simulation, we chose a time step after the re-connection event where the horseshoe structure is present. One characteristic of the horseshoe is that it appears as a weaker vortex in a plane orthogonal to the two parallel vortices. Another characteristic of interest, which applies to vortices in general, is that they have lower pressure at their cores. We therefore use these two ideas to design a trait that captures the horseshoe structure. Since the two parallel vortices are along the Z-axis and are separated from each other in the Y direction, we can expect the horseshoe to lie in the XY plane with the vortex core roughly aligned with the Y-axis. Considering the velocity vector field $\textbf{v}$ as a multi-field consisting of three velocity components $(v_x, v_y, v_z)$, we can formally convert the ideas above into a trait with high absolute values of $v_x$ and $v_z$ combined with simultaneously low absolute values of $v_y$. Further, we add the pressure criterion to the trait and set it below. Fig.~\ref{fig:flow} shows the results obtained for this trait. As evident from the feature level sets shown in Fig.~\ref{fig:flow-df}, we can see the horseshoe-like structure on the left within the green and purple level sets. These two level sets are also closest to our trait as evident from the histogram in Fig.~\ref{fig:flow-tf}.

For the final extraction of the horseshoe structure, we employ the branch decomposition segmentation derived from the trait-induced merge tree. As shown in Fig.~\ref{fig:flow-seg}, the purple segment captures the horseshoe structure. This result suggests feature level sets and trait-induced merge trees can be utilized for extracting features from multi-fields, which are difficult to extract otherwise, using simple queries. 

\section{Discussion}
\label{sec:discussion}
The method presented provides a novel approach of topological data analysis for multi-variate data and tensor fields and is capable of producing valuable insight into the data. 
However, in order to achieve these results, special attention needs to be paid in its implementation and usage.

Conceptually, the cutoff method for querying the merge tree should provide an intuitive segmentation of the domain. In practice, it proved to be hard to steer without visual aid. In order to find an adequate cutoff threshold, a persistence diagram or a tree representation with a cutoff slider could potentially make the method more usable and useful.
Regarding the distance field computation, the choice of distance metric plays a significant role for the resulting distance field and will therefore influence the tree computation. The advantage of the feature level-set method is that the distance metric can be exchanged for the most appropriate one. 
However, determining the trait will normally depend on domain knowledge.
As for the determination of the trait, this is a task that requires some domain knowledge.
%
%
Additionally, interpolation artifacts may arise for non-linear data such as tensor fields. In order to compute exact level sets, the data would need to be interpolated in the spatial domain during the ray-marching process instead of pre-computing the distance field. For a thorough discussion of these aspects of FLS we refer the reader to \cite{JankowaiHotz2019}.
%

%
%

%

Despite these challenges, the method presented is conceptually simple as opposed to previously proposed methods. The computation of feature level sets is a straightforward process of calculating a distance field and merge tree computation is both well-researched and available via open-source libraries.
This simple concept allows for the generalization of fundamental topological concepts with all their strength, as here demonstrated with the merge tree. Notions like persistence can be naturally extended to multi-fields or tensor fields supporting a multi-scale analysis.

On the user side, the interface consists of first defining the trait and then browsing the segmentation via a legend or slice. For a domain expert or educated user, defining the trait should be simple enough, given the nature of data analysis is targeted instead of exploratory. Once the trait is defined, the process is virtually automatic. Prior to rendering, an appropriate simplification threshold for the merge tree needs to be found in order to obtain sensible results for the segmentation. Visual aids such as persistence diagrams may be employed here. In the future, we plan to extend the supported rendering styles also including automatic transfer function design.

The implementation for FLS used in this article is GPU-based and performed at interactive frame rates for all data sets used. The computation of the merge tree took less than a minute for the selected data sets. Feature extraction based on the different methods presented in \sref{sec:query_methods} took a comparable amount of time. Additionally, the process of feature extraction can be relayed onto the GPU and should perform at interactive frame rates. Given these numbers, we argue that the application is usable in workflows for domain experts.

\section{Conclusions}
\label{sec:conclusions}
In this paper we have introduced topology-based segmentation of multi-field data based on traits.
As introduced in \cite{JankowaiHotz2019}, traits are geometries defined in attribute space describing parameters of interest. After the distance from this trait has been computed for every vertex or cell, the resulting distance field is pulled back into the spatial domain. Subsequently, we perform a segmentation of this distance field based on merge trees. Finally, we allow the user to query the merge tree with a set of different methods and parameters and generate according renderings. This approach provides a entirely new link to topological analysis of multi-fields and tensor fields.

For the three case studies presented in this article, the proposed method produced the expected and desired outcomes. In particular, the extraction of donor and acceptor regions yielded results greatly appreciated by the domain scientist. Hence, we hope the field of computational chemistry may benefit from this method. The investigation of automating the segmentation process is a promising extension of this work.

\acknowledgments{
This work is supported by \href{http://www.e-science.se/}{Swedish e-Science Research Centre (SeRC)}, \href{https://elliit.se/?l=en}{ELLIIT} environment for strategic research in Sweden, and the Swedish Research Council (VR) grant 2019-05487.
The application was implemented using the open-source software Inviwo \cite{inviwo2019}.
%
The computation of contour/merge trees uses code provided by \href{https://github.com/harishd10/contour-tree}{Harish Doraiswamy}.
The authors express their gratitude to Mathieu Linares for providing the simulation data for charge transfer and for providing very useful expert feedback on the obtained results and visualisations for this case study.
%
%
The flow data set used in this paper was produced and supplied by Professor Jan Nordstr\"om, Department of Mathematics, Link\"oping University and Dr. Marco Kupiainen, Rossby Centre, SMHI.





}

\bibliographystyle{abbrv-doi}

\bibliography{bibliographies/multifield, bibliographies/topology, bibliographies/acknowledgments, bibliographies/linsen}
\end{document}